\documentclass{article}
\usepackage{graphicx}
\usepackage{caption}
\usepackage[ruled,vlined]{algorithm2e}
\usepackage{xcolor, wrapfig,lipsum}
\usepackage{gensymb}
\usepackage{amsmath}
\usepackage{array, enumitem}
\usepackage{multirow}
\usepackage[font=small,labelfont=bf]{caption}
\usepackage[nonatbib, final]{neurips_2020}

\usepackage[utf8]{inputenc} % allow utf-8 input
\usepackage[T1]{fontenc}    % use 8-bit T1 fonts
\usepackage{hyperref}       % hyperlinks
\usepackage{url}            % simple URL typesetting
\usepackage{booktabs}       % professional-quality tables
\usepackage{amsfonts}       % blackboard math symbols
\usepackage{nicefrac}       % compact symbols for 1/2, etc.
\usepackage{microtype}      % microtypography

\title{Inductive Predictions of Extreme Hydrologic Events in The Wabash River Watershed}

\author{%
  Nicholas Majeske$^*$, Bidisha Abesh$^\dagger$, Chen Zhu$^\dagger$, Ariful Azad$^*$\\
  nmajeske@iu.edu, bidabesh@iu.edu, chenzhu@iu.edu, azad@iu.edu \\
  $^*$Department of Intelligent Systems Engineering, $^\dagger$Department of Earth and Atmospheric Sciences\\
  Indiana University Bloomington, IN, USA\\
}

\begin{document}
\vspace{-15pt}
\maketitle
\vspace{-25pt}
\begin{abstract}
\vspace{-8pt}
  We present a machine learning method to predict extreme hydrologic events from spatially and temporally varying hydrological and meteorological data.
We used a timestep reduction technique to reduce the computational and memory requirements and trained a bidirection LSTM network to predict soil water and stream flow from time series data observed and simulated over eighty years in the Wabash River Watershed.
We show that our simple model can be trained much faster than complex attention networks such as GeoMAN without sacrificing accuracy.  
Based on the predicted values of soil water and stream flow, we predict the occurrence and severity of extreme hydrologic events such as droughts.
We also demonstrate that extreme events can be predicted in geographical locations separate from locations observed during the training process.
%This spatially-inductive setting can be very useful to predict extreme events in other areas of the US using our model trained with the Wabash River Watershed data.
This spatially-inductive setting enables us to predict extreme events in other areas in the US and other
parts of the world using our model trained with the Wabash Basin data.

\end{abstract}

\vspace{-15pt}
\section{Introduction}
\vspace{-10pt}
Extreme hydrologic events such as floods and droughts are among the most severe weather-related disasters that cause widespread damage in agriculture, wildlife habitats, and human properties.
For example, the 2012 drought in the
continental United States resulted in an estimated \$30 billion in mostly agricultural losses~\cite{rippey2015us}.
Hence, predicting the occurrence, frequency, and severity of extreme hydrologic events is extremely important for societal well-beings~\cite{reichstein2019deep}.

In this paper, we use machine learning (ML) models to predict extreme hydrologic events related to droughts and floods. 
A drought is defined as prolonged periods of below-average water availability~\cite{dracup1980definition} and a flood is defined as a sudden increase of water level. 
The severity of droughts and floods can be quantitatively measured by soil water content and stream flows. 
Thus, predicted soil water (SW) and stream flow (SF) values can be used to accurately forecast 
future droughts and floods.
Here, we use bidirectional Long Short-term Memory (BLSTM) networks to predict SW and SF from historic SW, SF, precipitation, temperature, and other records. 
Based on the predicted values of SW and SF, we forecast the occurrence and severity of extreme events.

We used both observed and simulated data spanning over 84 years from the Wabash River Watershed located in Midwest USA.
%, which is the largest northern tributary of the Ohio River.
%The Wabash basin is sub-divided into 1276 subbasins for the purpose of data collection and simulation.
%where data is and for each subbasin, the dataset includes daily hydrological recordings from 1929 to 2013.
%For each of 1276 subbasins, we have 13 time series containing  hydrological records spanning over 84 years.  
%We developed a novel time reduction technique to reduce the number of time steps without sacrificing the accuracy. 
We trained BLSTM networks-- one for each subbasin-- to predict SW and SF from time reduced historic data. 
These simple BLSTM networks, when used with a time reduction technique, outperform autoreggressive methods such as ARIMA~\cite{box1970distribution} and VAR ~\cite{zivot2006vector} and perform comparably with complex models such as GeoMAN~\cite{liang2018geoman}. 
%However, our LSTM network can be trained much faster than GeoMAN, which is important for large-scale datasets considered here.
To this end, we considered two settings: (a) {\em spatially-transductive} learning where a BLSTM network is trained and tested on data from the same subbasin and (b) {\em spatially-inductive} learning where a BLSTM network is trained on data from one subbasin and tested on data from a different subbasin.
The BLSTM networks perform well for both settings. 
%In this paper, we used historic
%More interestingly, the spatially-inductive learning does perform well, especially when the corresponding subbasins are spatially related.  
%The latter finding indicates that the trained model can be used to predict extreme events from other areas in the US.

\vspace{-10pt}
\section{Materials and Methods}
\vspace{-8pt}
\subsection{Wabash River Watershed Time-Series}
\vspace{-5pt}
The Wabash River Watershed time-series consists of daily hydrological and meteorological records from 1929-01-01 to 2013-12-31 for a total $|T| = 31,046$ timesteps. The basin is subdivided into $|S| = 1,276$ distinct subbasins each of which record $|F| = 5$ features; 2 hydrological (stream flow and soil water content) and 3 meteorological (precipitation, minimum and maximum temperature). In our implementation and for the purposes of this paper, the complete time-series is a $3^{rd}$-order tensor of dimensions $|T| \cdot |S| \cdot |F|$. Hydrological features of the time-series were obtained from simulation using the Soil and Water Assessment Tool (SWAT) hydrological model \cite{arnold1998large,arnold1995continuous}. Meteorological features are observed records used in-part to parameterize SWAT and were obtained from the University of Notre Dame \cite{boryan2011monitoring}. 

\vspace{-10pt}
\subsection{Timestep Reduction}
\vspace{-7pt}
The daily timestep granularity of the Wabash River Watershed time-series is precise but computationally cumbersome and superfluous in modeling extreme events such as drought which span numerous timesteps. We apply timestep reduction to reduce timestep granularity and significantly decrease memory, computational, and modeling complexity with minimal cost to model accuracy. To introduce this algorithm, let $|T|$ be the timestep count of the original time-series, $w$ be reduction window size ( 7: weekly), and $s$ be reduction window stride. Timestep reduction is trivially applied using $s = w$ but we generalize this to any $s$ such that $s$ evenly divides $w$ allowing control over timestep distinctiveness (varying $s$) in addition to granularity (varying $w$).
The common sliding window method for extracting window samples requires timestep contiguity. To ensure contiguity persists through timestep reduction, we create $\frac{w}{s}$ reduced time-series channels. Each channel is created by applying the simple $s = w$ reduction at an offset unique to that channel defined $c_{\text{off}} = c_{id} \cdot s$ where $c_{id} \in \{0,1,...,\frac{w}{s}-1\}$. Using the sliding window method, we then extract input/output window samples from each channel separately and pool them to create the final window sample sets $X$ and $Y$. 
%Refer to algorithm \ref{Alg:TimestepReduction} for pseudo-code on timestep reduction.

% \begin{algorithm}[H]
% \SetAlgoLined
% \SetKwProg{Fn}{Function}{ is}{end}
% \Fn{ReduceTimesteps($T: Tensor(|T| \cdot |S| \cdot |F|)$,  $w: int$, $s: int$)}{
%     $T_{reduced} \gets Tensor(\frac{w}{s} \cdot \frac{|T|}{w} \cdot |F|)$ \;
%     \For{$c_{id}$ in range($\frac{w}{s}$)}{
%         $c_{offset} \gets s \cdot c_{id}$ \;
%         \For{$i$ in range($c_{offset},|T|,w$)}{
%             $T_{reduced}[c_{id},:,:] \gets ReduceMean(T[i:i+w,:,:]$, $dim$=$0$) \;
%         }
%     }
%     return $T_{reduced}$ \;
% }
%  \caption{Timestep Reduction}
%  \label{Alg:TimestepReduction}
% \end{algorithm}

\begin{wraptable}{r}{6.9cm}
\vspace{-18pt}
%\begin{table}[ht]
\scalebox{.85}{
    \begin{tabular}{|c|c||c|c|c|}
    \hline
    w & s & Model & Testing NRMSE & Operation \\
    & & Timesteps &  SW $\ \ \ \ \ \ \ \  \ \ \ \ \ $ SF & Ratio \\
    \hline
    \hline
    1 & 1 & $84\rightarrow28$ & 0.0831 $\ \ \ \ \ \ $ 0.0657 & 1 \\
    \hline
    7 & 1 & $12\rightarrow4$ & 0.0815  $\ \ \ \ \ \ $ 0.0682 & 1/7 \\
    \hline
    7 & 7 & $12\rightarrow4$ & 0.0887 $\ \ \ \ \ \ $ 0.0701 & 1/49 \\
    \hline
    14 & 1 & $6\rightarrow2$ & 0.0828 $\ \ \ \ \ \ $ 0.0711 & 1/14 \\
    \hline
%    14 & 2 & $6\rightarrow2$ & 0.0813 & 0.0736 & 1/28\\
%    \hline
%    14 & 7 & $6\rightarrow2$ & 0.0887 & 0.0773 & 1/98 \\
%    \hline
    14 & 14 & $6\rightarrow2$ & 0.0917 $\ \ \ \ \ \ $ 0.0780 & 1/196 \\
    \hline
    28 & 1 & $3\rightarrow1$ & 0.0801 $\ \ \ \ \ \ $ 0.0806 & 1/28 \\
    \hline
%    28 & 2 & $3\rightarrow1$ & 0.0807 & 0.0807 & 1/54 \\
%    \hline
%    28 & 4 & $3\rightarrow1$ & 0.0808 & 0.0805 & 1/112 \\
%    \hline
%    28 & 7 & $3\rightarrow1$ & 0.0793 & 0.0811 & 1/196 \\
%    \hline
%    28 & 14 & $3\rightarrow1$ & 0.0802 & 0.0881 & 1/392 \\
%    \hline
    28 & 28 & $3\rightarrow1$ & 0.0889 $\ \ \ \ \ \ $ 0.0894 & 1/784 \\
    \hline
    \end{tabular}
    }
    \vspace{-5pt}
    \caption{Effect of parameters $w$ and $s$ on model generalization and computational complexity.}
    \vspace{-15pt}
    \label{Table:TimestepReductionParameterEffectOnNRMSE}
%\end{table}
\end{wraptable} 
Table \ref{Table:TimestepReductionParameterEffectOnNRMSE} briefly analyzes the effect of $w$ and $s$ on computational complexity and model generalization measured by testing set normalized root mean squared error (NRMSE) defined equation in \eqref{Equation:NRMSE}. We find increasing $w$ decreases epoch operation count at the cost of window sample count and decreasing $s$ reduces window sample count loss at the cost of window sample distinctiveness. Ultimately, timestep reduction to weekly units with $w = 7$ and stride $s = 1$ provides an optimal balance between computational complexity and model generalization.
% \begin{table}[!h]
%     \centering
%     \captionsetup{justification=centering,margin=2cm}
%     %\captionsetup{justification=centering,margin=2cm}
%     \begin{tabular}{|c|c||c|c|c|c|}
%     \hline
%     w & s & Model  & SW & SF & Operation \\
%      & & Timesteps & & & Ratio \\
%     \hline
%     \hline
%     1 & 1 & $84\rightarrow28$ & 0.0831 & 0.0657 & 1 \\
%     \hline
%     7 & 1 & $12\rightarrow4$ & 0.0815 & 0.0682 & 1/7 \\
%     \hline
%     7 & 7 & $12\rightarrow4$ & 0.0887 & .0701 & 1/49 \\
%     \hline
%     14 & 1 & $6\rightarrow2$ & 0.0828 & 0.0711 & 1/14 \\
%     \hline
% %    14 & 2 & $6\rightarrow2$ & 0.0813 & 0.0736 & 1/28\\
% %    \hline
% %    14 & 7 & $6\rightarrow2$ & 0.0887 & 0.0773 & 1/98 \\
% %    \hline
%     14 & 14 & $6\rightarrow2$ & 0.0917 & 0.0780 & 1/196 \\
%     \hline
%     28 & 1 & $3\rightarrow1$ & 0.0801 & 0.0806 & 1/28 \\
%     \hline
% %    28 & 2 & $3\rightarrow1$ & 0.0807 & 0.0807 & 1/54 \\
% %    \hline
% %    28 & 4 & $3\rightarrow1$ & 0.0808 & 0.0805 & 1/112 \\
% %    \hline
% %    28 & 7 & $3\rightarrow1$ & 0.0793 & 0.0811 & 1/196 \\
% %    \hline
% %    28 & 14 & $3\rightarrow1$ & 0.0802 & 0.0881 & 1/392 \\
% %    \hline
%     28 & 28 & $3\rightarrow1$ & 0.0889 & 0.0894 & 1/784 \\
%     \hline
%     \end{tabular}
%     \caption{Effect of timestep reduction parameters $w$ and $s$ on soil water and stream flow generalization and computational complexity.}
%     \label{Table:TimestepReductionParameterEffectOnNRMSE}
% \end{table}

\begin{wrapfigure}{r}{7.2cm}
\vspace{-30pt}
%\begin{figure}[!h]
    \centering
    %\captionsetup{justification=centering}
    \includegraphics[width=0.49\textwidth]{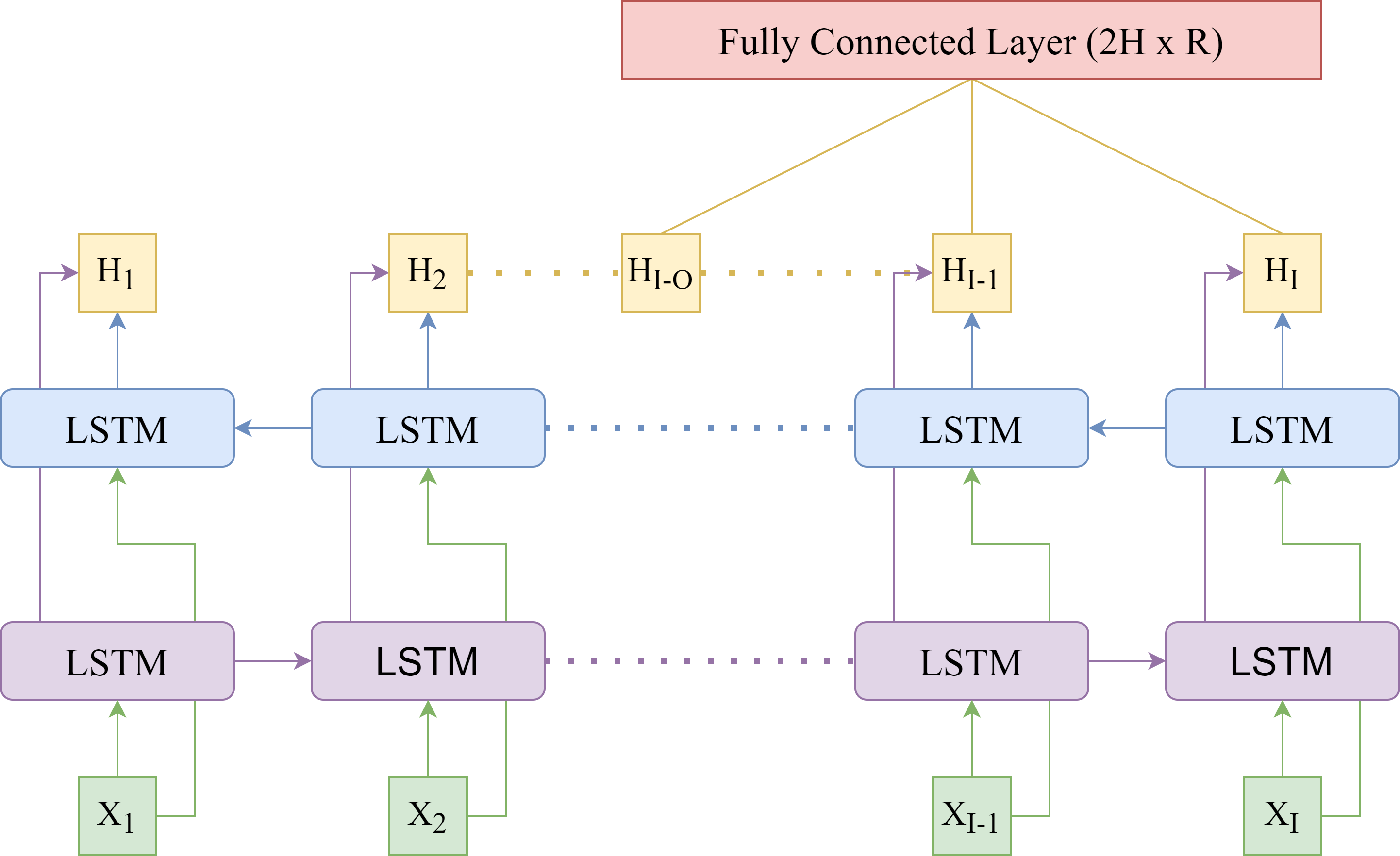}
    \vspace{-12pt}
    \caption{The Bidirectional LSTM network architecture.}
    \vspace{-10pt}
    \label{Figure:BLSTM}
%\end{figure}
\end{wrapfigure}

\vspace{-8pt}
\subsection{Feature Standardization}
\vspace{-5pt}
To facilitate model optimization, we standardize all predictor and response features. 
The Wabash River Watershed time-series presents an interesting challenge for standardization since mean and standard deviation vary across both temporal ($T$) and spatial ($S$) domains. 
Across $T$, we see a cyclical change in mean and standard deviation coinciding with seasonality. Across $S$ we see a wide range of mean and standard deviations coinciding with low (potentially upstream) and high (potentially downstream) yield subbasins. In order to correctly standardize our features we calculate mean and standard deviation specific to the day of year and subbasin resulting in $3^{rd}$-order tensors of dimensions $366 \cdot |S| \cdot |F|$.

\vspace{-10pt}
\subsection{Bidirectional LSTM Networks}
\vspace{-5pt}
The LSTM network proposed in \cite{hochreiter1997long} receives as input a series of $I$ timesteps each containing $P$ predictors and recurrently transforms them into a set of $I$ hidden states $H_{1},...,H_{I}$ each containing $h$ hidden units. Each hidden state $H_{t}$ encodes information observed in all previous and current inputs $X_{1},...,X_{t}$ and all previous hidden states $H_{1},...,H_{t-1}$. A significant limitation of this architecture is the constraint of any hidden state $H_{t}$ to encode only information observed in previous and current timesteps $1,...,t$ while valuable predictor information for timestep $t$ may exist in future timesteps $t+1,...,I$. Specifically, if observing $I$ input timesteps and predicting $O$ output timesteps, this architecture is constrained to use only the first $I-O+t$ timesteps when predict timestep $t$ of the output series. For non-trivial timestep mappings ($I \geq O > 1$ and $I - O = \epsilon$) a significant $I - t$ timesteps are omitted in the prediction of timestep $t$ though these timesteps share greater temporal proximity with timestep $t$. For this reason we use bidirectional LSTM networks which remove the constraint by processing information both forwards and backwards through time. Thus, hidden state $H_{t}$ may now encode information from all input timesteps $X_{1},...,X_{t},...X_{I}$ and all other hidden states $H_{1},...,H_{t-1},H_{t+1},...,H_{I}$.

\vspace{-10pt}
\section{Results}
\vspace{-10pt}
\subsection{Experiment Settings}
\vspace{-7pt}
We implemented our BLSTM in PyTorch 1.5.0~\cite{PyTorch} and  executed all experiments on an Acer Predator G3-710 equipped with an Intel Core i7-7700 at 3.6GHz, 32GB RAM, and Nvidia's GeForce 1070 (1920 CUDA cores, 1.5GHz, 8GB memory). The entire time-series was split into training, validation, and testing sets containing all timesteps over 1929/10/01-1997/09/30, 1997/10/01-2005/09/30, and 2005/10/01-2013/09/30 respectively. To further evaluate the efficacy of our model, we established 3 baselines for comparison:
\begin{itemize}[topsep=-3pt, itemsep=-2pt, leftmargin=15pt]
    \item \textbf{Naive Last-Timestep}: A non-parametric model predicting for all timesteps $O$ observation $X_{I}$.
    \item \textbf{ARIMA} \cite{box1970distribution}: A classical auto-regressive model for time-series forecasting.
    \item \textbf{GeoMAN} \cite{liang2018geoman}: A complex LSTM network with temporal and spatial attention.
\end{itemize}

We evaluated both the spatially transductive and inductive capability of our BLSTM and all 3 baselines for subbasin pairs $\{s_{trans},s_{ind}\} \in \{\{1,2\},\{141,142\},\{1275,1276\}\}$. Each model was optimized to the training set of the spatially transductive subbasin $s_{trans} \in \{1,141,1275\}$ saving the model state with minimum $MSE$ loss over validation set. Our best testing set NRMSE for BLSTM ($I = 12$ weeks, $O = 4$ weeks, $R = |\{Soil Water, Stream Flow\}| = 2$) was acheived using $P = 6$ predictors including \{Date, Minimum Temperature, Maximum Temperature, Precipitation, Soil Water, Stream Flow\}, $h = 256$ hidden units, mini-batch stochastic gradient descent (mini-batch size 128), learning rate 0.01 (and learning rate decay 0.001), 0.001 $l_{2}$ regularization, Xavier initialization, and 50 epochs of training.

\vspace{-10pt}
\newcommand{\AddHat}[1]{\expandafter\hat#1}
\begin{equation}
\text{NRMSE} = \frac{\sqrt{E(\AddHat{y} - y)^{2})}}{y_{max} - y_{min}} = \frac{\text{RMSE}}{y_{max} - y_{min}}
\label{Equation:NRMSE}
\end{equation}
% NRMSE = \frac{\sqrt{\frac{\sum_{i=1}^{|T|}(\AddHat{y_i} - y_i)^{2}}{|T|}}}{y_{max} - y_{min}}

\begin{table}[!t]
    \centering
    %\captionsetup{justification=centering,margin=2cm}
    \scalebox{.88}{
    \begin{tabular}{|c|c c|c c c |c c c|}
    \hline
     & \multicolumn{2}{c|}{Training} &  \multicolumn{3}{c|}{Testing (spatially transductive)} & \multicolumn{3}{c|}{Testing (spatially inductive)} \\
    Model & Subbasin & Time (sec) & Subbasin & SW & SF & Subbasin & SW & SF \\
    \hline
    \hline
    Naive Last-Timestep & 1  & N/A & 1 & 0.0950 & 0.0867 & 2 & 0.0783 & 0.1898\\
    ARIMA & 1 & 0.58 & 1 & 0.2201 & 0.0871 & 2 & 0.1898 & 0.1968 \\
    GeoMAN & 1 & 232.6 & 1 & 0.0815 & 0.0677 & 2 & 0.0734 & 0.1565 \\
    BLSTM & 1 & 51.4 & 1 & 0.0815 & 0.0682 & 2 & 0.0768 & 0.1599 \\
    \hline
    \hline
    Naive Last-Timestep & 141 & N/A & 141 & 0.0920 & 0.1275 & 142 & 0.0916 & 0.1092 \\
    ARIMA & 141 & 0.72 & 141 & 0.2060 & 0.1272 & 142 & 0.2000 & 0.1063  \\
    GeoMAN & 141 & 230.5 & 141 & 0.0787 & 0.0961 & 142 & 0.0803 & 0.0814 \\
    BLSTM & 141 & 51.7 & 141 & 0.0806 & 0.0985 & 142 & 0.0810 & 0.0845  \\
    \hline
    \hline
    Naive Last-Timestep & 1275 & N/A & 1275 & 0.0977 & 0.1220 & 1276 & 0.1040 & 0.1240 \\
    ARIMA & 1275 & 0.53 & 1275 & 0.1746 & 0.1786 & 1276 & 0.1942 & 0.1322 \\
    GeoMAN & 1275 & 231.4 & 1275 & 0.0908 & 0.0842 & 1276 & 0.0892 & 0.1202 \\
    BLSTM & 1275  & 52.1 & 1275 & 0.0808 & 0.0849 & 1276 & 0.0860 & 0.1435\\
    \hline
    \end{tabular}
    }
    %\newline
    \caption{Model performances (normalized RMSE scores) when predicting soil water and stream flow.}
    \vspace{-18pt}
    \label{Table:SpatiallyTansductiveAndInductiveNRMSEs}
\end{table}

\vspace{-10pt}
\subsection{Transductive and Inductive Predictions of Soil Water}
\vspace{-8pt}
Table \ref{Table:SpatiallyTansductiveAndInductiveNRMSEs} shows testing set NRMSE on SW for our BLSTM and all 3 baselines. 
%An important note to make is that Naive Last-Timestep performs simple static prediction and we should expect the 3 other dynamical models to outperform this baseline. 
%However, 
ARIMA performs significantly worse in predicting SW. Upon further investigation of the plotted predictions, ARIMA is predicting for 4 weeks an approximate mean over the 12 observed weeks. SW varies significantly at the week-to-week level causing this type of prediction to systematically over and under project. In all cases, GeoMAN and BLSTM outperform Naive Last-Timestep but we see either comparable NRMSE between GeoMAN and BLSTM or one model outperform the other. GeoMAN outperforms BLSTM for subbasin pairs $\{1,2\}$ and $\{141,142\}$ but BLSTM significantly outperforms GeoMAN for subbasin pair $\{1275,1276\}$. The otherwise even performance between GeoMAN and BLSTM is interesting considering the discrepancy in model complexity. This is an advantage to BLSTM, which sees a
%$\frac{231.5}{51.7} = 4.5\times$ 
$4.5\times$ speed-up in training runtime over GeoMAN given an identical training regime.

Moving from spatially transductive to inductive prediction shows either reduced, approximately unchanged, or increased NRMSE. For subbasin pair $\{1,2\}$ spatial induction outperforms transduction while performance is approximately equal in subbasin pair $\{141,142\}$. Performance is somewhat degraded when moving to inductive prediction for subbasin pair $\{1275,1276\}$ for BLSTM but marginally improved for GeoMAN. The left plot of Fig. \ref{fig:SoilWaterAndStreamFlowInferenceAndGroundthCurves} shows spatially transductive and inductive SW predictions for subbasin pair $\{1,2\}$ using BLSTM, where both settings accurately predict ground truth values.

\vspace{-5pt}
\subsection{Transductive and Inductive Predictions of Stream Flow}
\begin{figure}[!t]
    \centering
    %\captionsetup{justification=centering,margin=2cm}
    \vspace{-5pt}
    \includegraphics[width=1.03\textwidth]{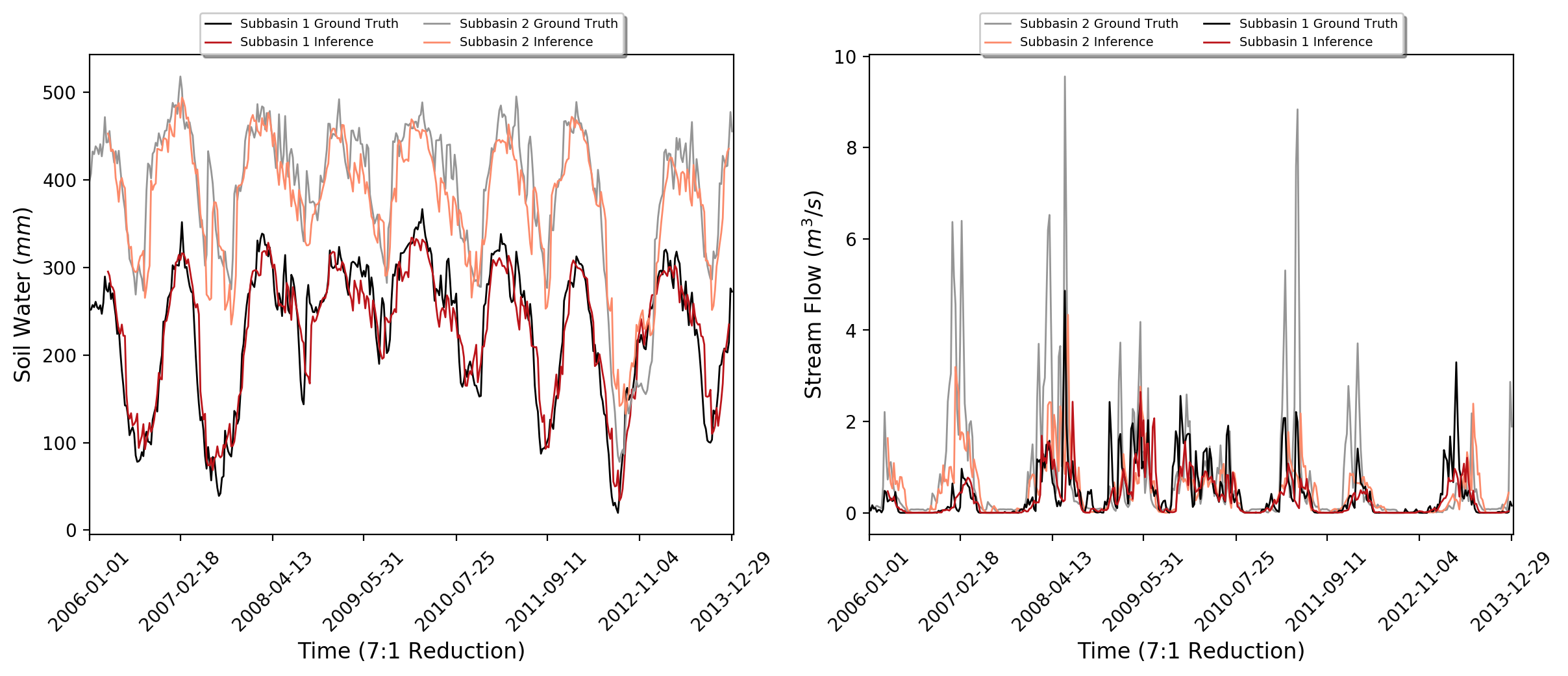}
    \vspace{-17pt}
    \caption{Testing set inference for spatially transductive and inductive settings on (left) soil water and (right) stream flow using subbasins 1 and 2.}
    \vspace{-10pt}
    \label{fig:SoilWaterAndStreamFlowInferenceAndGroundthCurves}
\end{figure}

\vspace{-5pt}
Table \ref{Table:SpatiallyTansductiveAndInductiveNRMSEs} additionally shows testing set NRMSE on SF for our BLSTM and all 3 baselines. ARIMA performs on-par with Naive Last-Timestep for all subbasin pairs with the exception of $\{1275,1276\}$ in which performance is significantly worse. GeoMAN and BLSTM show approximately equal performance for spatially transductive subbasins but GeoMAN systematically outperforms BLSTM for spatially inductive subbasins. This is likely due to the superior complexity of GeoMAN relative to BLSTM allowing for better fit to the high frequency of extreme values we observe in SF. The right plot of Figure \ref{fig:SoilWaterAndStreamFlowInferenceAndGroundthCurves} overlays spatially transductive and inductive SF predictions for subbasin pair $\{1,2\}$.
This plot confirms BLSTM's trouble in predicting extreme values in SF which accounts for the higher NRMSE we observe.

\vspace{-10pt}
\subsection{Extreme Event Prediction}
\vspace{-5pt}
Each timestep of the ground truth and predicted curves are classified according to a standardized index simplified from the drought index presented in \cite{dierauer2020drought}. Specifically, each predicted timestep is converted to its z-score $z$ and classified based on the following intervals:
Normal: $z \in [-1,1]$, Moderate: $z \in [-1.5,-1) \ or \ z \in (1,1.5]$, Severe: $z \in [-2,-1.5) \ or \ z \in (1.5,2]$, and Extreme: $z \in (-\infty,-2) \ or \ z \in (2, \infty)$.
    
% \begin{itemize}
%     \item Normal (N): $z \in [-1,1]$
%     \item Moderate (M): $z \in [-1.5,-1) \ or \ z \in (1,1.5]$
%     \item Severe (S): $z \in [-2,-1.5) \ or \ z \in (1.5,2]$
%     \item Extreme (E): $z \in (-\infty,-2) \ or \ z \in (2, \infty)$
% \end{itemize}

\begin{wrapfigure}{r}{8.4cm}
\vspace{-24pt}
%\begin{figure}[!h]
    \centering
    \captionsetup{justification=centering}
    \includegraphics[width=0.57\textwidth]{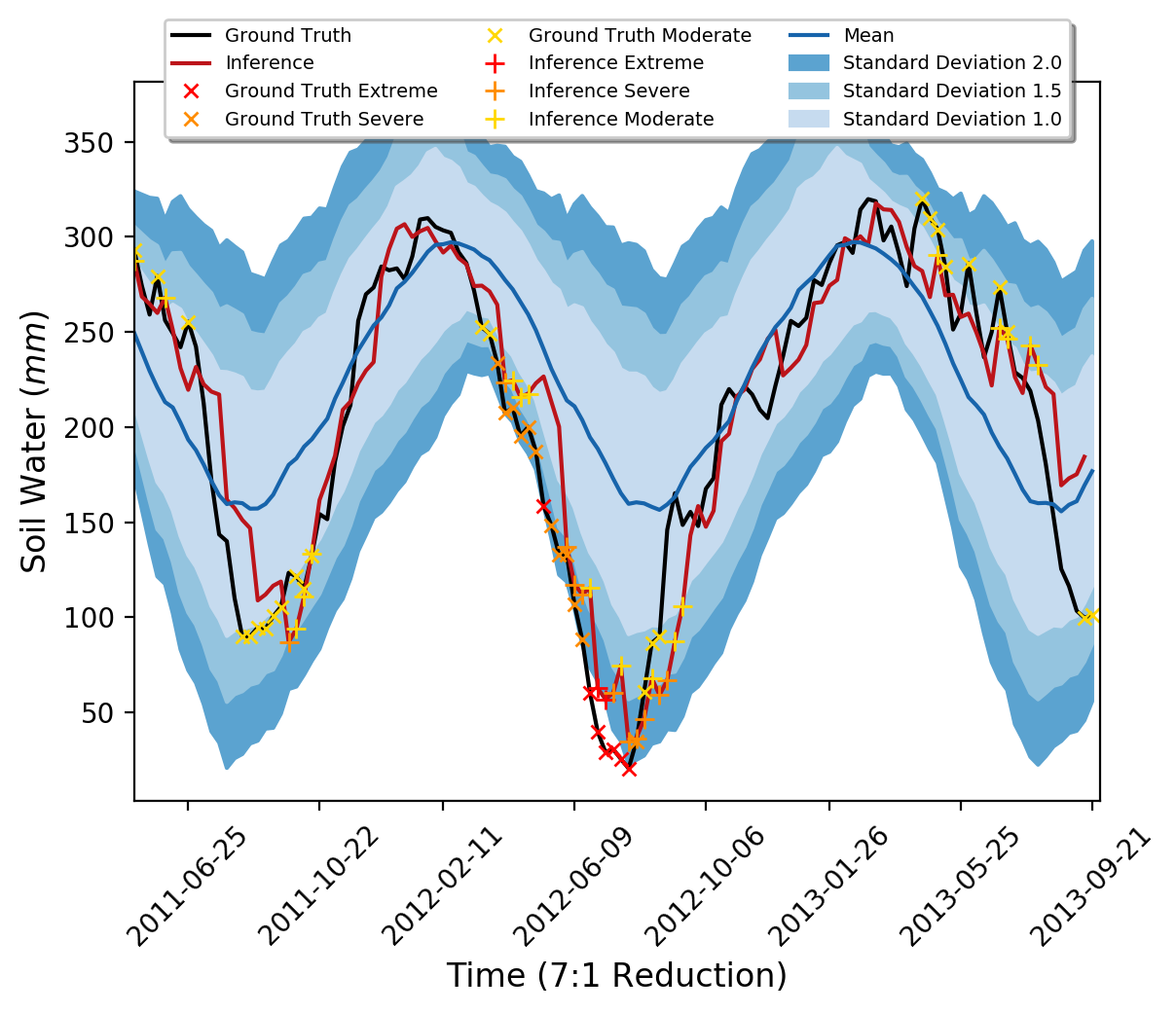}
    \vspace{-15pt}
    \caption{ Extreme event classification on soil water for subbasin 1.}
    \vspace{-15pt}
    \label{fig:SoilWaterEventMarkers}
%\end{figure}
\end{wrapfigure}
Fig. \ref{fig:SoilWaterEventMarkers} shows a subsection of the testing set for subbasin 1 with extreme events labeled for ground truth and prediction on soil water. The blue regions enclose normal, moderate, and severe intervals with extreme events ($|z| > 2$) falling outside of all regions. Currently, our BLSTM tends to under project soil water values causing missclassification of some events to less extreme events. 
%To automate the classification, we trained a logistic regression model with z-scores of predicted SW values at each test timestep of subbasin 1.
%The right plot of Fig. \ref{fig:SoilWaterEventMarkers} shows the confusion matrix when extreme events are predicted on subbasin 2 using the trained logistic regression model.
%The confusion matrix confirms that BLSTM tends to cause missclassification of some events to less extreme events. 
This phenomena is even more prevalent in stream flow prediction where extreme values are more frequent and of greater magnitude.

% \begin{figure}[!t]
%     \centering
%     \vspace{-1pt}
%     %\captionsetup{justification=centering,margin=2cm}
%     \includegraphics[width=0.48\textwidth]{Figures/EventMarkers_Subbasins[1]_Responses[SWmm,FLOW_OUTcms]_2.png}
%     %\includegraphics[width=0.48\textwidth]{Figures/confusion.png}
%     \vspace{-8pt}
%     \caption{(left) Extreme event classification on soil water. %(right) The confusion matrix when extreme events in subbasin 2 are predicted from a logistic regression model trained with extreme events in subbasin 1.
%     }
%     \vspace{-12pt}
%     \label{fig:SoilWaterEventMarkers}
% \end{figure}

\vspace{-8pt}
\section{Concluding Remarks}
\vspace{-10pt}
In this paper, we apply BLSTM networks and 3 baseline models to predict soil water and stream flow in both spatially transductive and inductive settings. We apply timestep reduction to decrease memory, computational, and model complexity and train a BLSTM to comparable generalization w.r.t. the highly complex GeoMAN \cite{liang2018geoman} at a fraction of the runtime. Using our soil water and stream flow predictions, we additionally perform extreme event classification for each timestep. Our experiments suggest (1) a relatively simple model can perform comparably well against complex models, (2) spatial induction can be performed with reliable accuracy for soil water but is currently unreliable for stream flow and (3) extreme event classification can be done with reasonable accuracy. In future work we plan to incorporate spatial information into the BLSTM architecture and scale the number of training subbasins. To handle the increased memory, computational, and model complexity we will also implement batch parallelism as described in \cite{gholami2018integrated}.
%\section*{References}
\bibliographystyle{abbrv}
\bibliography{main}
% References follow the acknowledgments. Use unnumbered first-level heading for
% the references. Any choice of citation style is acceptable as long as you are
% consistent. It is permissible to reduce the font size to \verb+small+ (9 point)
% when listing the references.
% {\bf Note that the Reference section does not count towards the eight pages of content that are allowed.}
% \medskip

% \small

% [1] Alexander, J.A.\ \& Mozer, M.C.\ (1995) Template-based algorithms for
% connectionist rule extraction. In G.\ Tesauro, D.S.\ Touretzky and T.K.\ Leen
% (eds.), {\it Advances in Neural Information Processing Systems 7},
% pp.\ 609--616. Cambridge, MA: MIT Press.

% [2] Bower, J.M.\ \& Beeman, D.\ (1995) {\it The Book of GENESIS: Exploring
%   Realistic Neural Models with the GEneral NEural SImulation System.}  New York:
% TELOS/Springer--Verlag.

% [3] Hasselmo, M.E., Schnell, E.\ \& Barkai, E.\ (1995) Dynamics of learning and
% recall at excitatory recurrent synapses and cholinergic modulation in rat
% hippocampal region CA3. {\it Journal of Neuroscience} {\bf 15}(7):5249-5262.

% [4] Boryan, C.; Yang, Z.; Mueller, R.; Craig, M. Monitoring US Agriculture: The US Department of Agriculture, National Agricultural Statistics Service Cropland Data Layer Program. Geocarto Int. 2011, 26, 341–358.

\end{document}